\documentclass[review]{elsarticle}

\usepackage{lineno,hyperref}

\journal{Elsarticle }

\usepackage{booktabs}
\usepackage{multirow}
\usepackage{amsmath}
\usepackage{fixmath}
\usepackage[normalem]{ulem}
\useunder{\uline}{\ul}{}
\graphicspath{{fig/}}
\usepackage{xcolor}
\usepackage{amsfonts}
\usepackage{geometry}

\usepackage{color}
\usepackage{graphicx}
\usepackage{subfig}
\usepackage{epsfig}
\usepackage{amsmath}
\usepackage{enumerate}
\usepackage{subfig}

\usepackage[linesnumbered,ruled,vlined]{algorithm2e}
\usepackage{comment}
\usepackage{multirow}
\usepackage{dcolumn}
\newcolumntype{d}[1]{D{.}{.}{#1}}
\begin{document}
	
	\begin{frontmatter}
		
		\title{  Deep Learning based Monocular Depth Prediction: Datasets, Methods and Applications}

		\author[firstaddress]{Qing Li}
		\author[firstaddress,secondaryaddress,thirdaddress,fifthaddress]{Jiasong Zhu \corref{correspondingauthor}}
		\cortext[correspondingauthor]{Corresponding author}
		\ead{zhujiasong@gmail.com}
		
		\author[sixth]{Jun Liu}
		\author[secondaryaddress,thirdaddress]{Rui Cao}
		
		\author[firstaddress,secondaryaddress,thirdaddress]{Qingquan Li}
		\author[forthaddress,fifthaddress]{Sen Jia}
		\author[seventh,eighth,ninth]{Guoping Qiu}

		
		\address[firstaddress]{College of Civil and Transportation Engineering, Shenzhen University, Shenzhen 518060, China}
		\address[secondaryaddress]{MNR Key Laboratory for Geo-Environmental Monitoring of Great Bay Area, Shenzhen University, Shenzhen 518060, China}
		\address[thirdaddress]{Guangdong Key Laboratory of Urban Informatics, Shenzhen University, Shenzhen 518060, China}
		\address[forthaddress]{College of Computer Science and Software Engineering, Shenzhen University, Shenzhen 518060, China}
		\address[fifthaddress]{Institute of Urban Smart Transportation and Safety Maintenance, Shenzhen University, Shenzhen 518060, China}
		\address[sixth]{Shenzhen TCL Industrial Research Institute Co., Ltd., Shenzhen, China}
		
		\address[seventh]{College of Electronics and Information Engineering, Shenzhen University, Shenzhen 518060, China}
		\address[eighth]{Guangdong Key Laboratory of Intelligent Information Processing, Shenzhen University, Shenzhen 518060, China}
		\address[ninth]{School of Computer Science, University of Nottingham, Nottingham NG8 1BB, UK}

		\begin{abstract}
			Estimating depth from RGB images can facilitate many photometric computer vision tasks, such as indoor localization, height estimation, and simultaneous localization and mapping (SLAM). Recently, monocular depth estimation has obtained great progress owing to the rapid development of deep learning techniques. They surpass traditional machine learning-based methods by a large margin in terms of accuracy and speed. Despite the rapid progress in this topic, there are lacking of a comprehensive review, which is needed to summarize the current progress and provide the future directions. In this survey, we first introduce the datasets for depth estimation, and then give a comprehensive introduction of the methods from three perspectives: supervised learning-based methods, unsupervised learning-based methods, and sparse samples guidance-based methods. In addition, downstream applications that benefit from the progress have also been illustrated. Finally, we point out the future directions and conclude the paper.
		\end{abstract}
		
		
	\end{frontmatter}

	\section{Introduction}\label{sec:introduction}
	Depth estimation from monocular images is an important research topic in the photometric computer vision community. It aims to produce pixel-wise depth maps from images at certain view perspectives. Such depth information helps better understand  3D scenes and also facilitates many computer vision tasks such as  indoor localization \cite{li20203d}, height estimation \cite{amirkolaee2019height}, simultaneous localization and mapping (SLAM) \cite{tateno2017cnn}, visual odometry \cite{yang2018deep}, classification \cite{he2017estimated}, etc. Usually, the depth information is obtained through commercial depth sensors such as various LiDAR devices and Kinects. However, apart from the high cost and requirement of operational skill, they also confront the disadvantages of low resolution and short perception distance, which limit extensive applications. Inferring depth map from monocular images attracts increasing interests due to the wide availability of RGB images. However, it is a challenging ill-posed problem as it is inherently scale ambiguous, which means that an infinite number of possible depth maps can be associated with an image. 
	
	Depth cues in monocular images have been exploited as hints for depth estimation. Guo et al. leverage the hazing generation principle for depth estimation \cite{guo2014adaptive}. They simulate a hazed image by adding a haze veil on the input image and use a haze removal algorithm to estimate the depth map. Imaging principle are exploited for depth estimation in \cite{tang2015depth}, where scene at the focus distance are clear while the others are blur. They first predict depth at the edge positions and recover the full depth map with depth matting optimization. Occlusion reasoning and vanish lines and points are leveraged to generate depth map indirectly in \cite{hoiem2007recovering,tsai2006block-based}. Though these methods have been proven to be effective in their experiments, they are not widely applied since the depth cues are not always presented in normal images.
	
	Another direction for depth estimation is to utilize machine learning techniques. They can be categorized into two groups: non-parametric methods and learning-based methods. Non-parametric methods exploit vast available RGBD datasets and infer the target depth from the multiple reference depth images. The reference depths are determined according to the image similarity of RGB domain between targeted image and the rest images. Konrad et al. \cite{konrad2013learning} search $K$ candidate depth image based on the histograms of oriented features. They generate the initial depth with median operator and refine it with cross-bilateral filtering. GIST feature and optical flow feature are exploited to find $K$ similar images and produce the initial depth map through a warping procedure based on SIFT flow \cite{karsch2014depth}. A global optimization is conducted to smooth and interpolate the warped initial depth. Herrera et al. \cite{herrera2014learning} retrieve candidate depth images using LBP feature and the preliminary depth is produced through weighted combination of depth candidates and refined with the joint bilateral filtering to enforce the depth consistency. Structure information is believed to be highly related with depth information in \cite{herrera2018automatic}. Besides, the candidate depth images are found based on structure similarity in colour domain. The prior depth map is obtained through the weighted average fusion with retrieved candidates. The final depth map is optimized with segmentation guided filtering. Kong et al. \cite{kong2015intrinsic} also retrieve $K$ candidate video clips with GIST features and generate the preliminary depth map through sift flow warping. They finally enhance the depth boundary and contours using albedo and shading as well as the regularizer of output structure from motion. Choi et al. \cite{choi2015depth} retrieve $K$ RGB-D image pairs and generate $K$ depth gradient samples by constructing the correspondences between retrieved images. The initial depth map is generated through Poisson equations. The refined depth map is obtained using edge-aware median filter. Xu et al. \cite{xu2019depth} exploit the label transfer in gradient domain. They fused the candidate depth gradients through confidence analysis to generate the initial gradient depth map and refine it based on image boundaries. Finally, the depth map is constructed through Poisson function and smoothed with edge-aware filter. Non-parametric methods is of low efficiency since they have to search candidates images from large image databases.
	
	Learning-based methods directly estimate the depth value from various features and usually incorporate with graph-based models. Saxena et al. \cite{saxena2005learning} directly predict the depth values from multi-scale local feature and global feature, and depth relations between different locations are modelled with Markov random field. They further refine the approach with regions under the assumption that each superpixel corresponds to the single value. Liu et al. \cite{liu2010single} incorporate semantic information with local pixel appearance for supervised depth prediction. Ladický et al. \cite{ladick2014pulling} also predict the depth with semantic and visual features and they simplify the classifiers by introducing the constraints of perspective geometry. Cao et al. \cite{cao2010a} utilize an iterative framework for depth estimation and image segmentation. They first generate an initial image segmentation results and then utilize a multi-scale Markov random field to infer depth. Given the initial depth map, they refine the segmentation results based on a graph-based merging algorithm and the final depth is refined with the new segmentation map. Zhuo et al. \cite{zhuo2015indoor} estimate the depth from the local and global structure of the scenes and fuse the feature in a continuous conditional random field (CRF) for the depth prediction. Liu et al. \cite{liu2014discrete} cast depth prediction as a discrete-continuous problem. Depth in the same superpixels are treated as continuous while depth between superpixels are believed to discrete. 
	
	Inspired by the powerful performance of deep learning on image classification, object detection, and semantic segmentation, many researchers attempt to tackle monocular depth prediction with deep learning techniques. Various network architecture based on convolutional neural network (CNN) \cite{liu2010single}, recurrent neural network (RNN) \cite{wang2019recurrent} and generative adversarial network (GAN) \cite{almalioglu2019ganvo:} have been proposed to exploit RGB images. Superior performances have been  achieved compared with traditional methods. However, comprehensive survey is lacking to summarize the current progress and provide some hints to future directions. To bridge the gap, in this paper, we overview the monocular depth prediction approaches based on deep learning from four aspects: benchmark datasets, supervised methods, unsupervised methods, and methods with sparse depth points guidance. We also discuss the future direction and trends. 
	
	The rest paper is organized as follows. Section \ref{sec:dataset} introduces the widely used indoor and outdoor benchmarks and the associated evaluation metrics in monocular depth estimation. Section \ref{sec:sup} and Section \ref{sec:unsup} review the supervised and unsupervised methods for depth prediction respectively. We also review recent works in depth prediction with sparse points in section \ref{sec:sparse}. Current challenges and promising directions are discussed in Section \ref{sec:fut}. Finally, Section \ref{sec:con} concludes the paper.

	\section{Datasets} \label{sec:dataset}
	
	To train or evaluate the deep learning-based methods, RGBD image pairs are needed. Generally, any RGBD datasets can be used. The depth maps can be collected in various manners including RGBD cameras \cite{silberman_indoor_2012,chang_matterport3d_2017,dai_scannet_2017,armeni_joint_2017}, Laser scanners \cite{geiger_are_2012,saxena_make3d_2009,su_bayesian_2017,koch_evaluation_2018,strecha_benchmarking_2008} or multi-view stereo (MVS) approaches \cite{li_megadepth_2018,chen_single-image_2016,xian_monocular_2018}. Some datasets even use synthetic depth map \cite{mccormac_scenenet_2017,song_semantic_2017} rendering from virtual scenes.  A detailed dataset introduction can be found in \cite{koch2020comparison}. In this section, we illustrate  some widely used benchmarks and group them into ground images and remote sensing images benchmarks. They are typical indoor scenes, outdoors and hybrid scenes, respectively. Besides, the corresponding evaluation metrics are also elaborated.
	
	\subsection{Ground image benchmarks}
	\textbf{NYU Depth V2.}
	The NYU Depth V2 \cite{silberman_indoor_2012}   dataset is an indoor RGBD dataset collected primarily for scene understanding. It captured  464 various scenes from different buildings with a Microsoft Kinect, and  249 scenes are used for training and 215 scenes for testing. The dataset consists of 1449 image pairs, of which 795 images are used for training and the remaining 654 images for testing for depth estimation , which is a sub-dataset of NYU with segmentation label. The resolution of the images is $640\times 480$. The depth image are in-painted to fill the hole in depth images and aligned to the RGB image. For sparse sample-based approaches, the sparse depth map are generated by choosing depth points from the depth map.
	
	\textbf{KITTI.}
	KITTI \cite{geiger_are_2012} dataset is an outdoor dataset collected  with a mobile mapping vehicles. The RGB images are captured with a stereo calibrated and rectified camera. The depth images are obtained by a rotating Velodyne laser scanner mounted on a driving vehicle. The ground-truth depth maps are generated by projecting the 3D points from the LiDAR laser into the left RGB camera with the given intrinsic and extrinsic parameters. Since the 3D points are not dense enough, the generated images are very sparse. The  size of the image is about $1224\times 368$. They are usually cropped to reduce the sky regions. Since their are collected with stereo cameras in continuous manner, this dataset is feasible to support stereo-based unsupervised methods and video-based unsupervised methods. For depth completion tasks, the KITTI dataset contains over 93,000 sparse depth maps and original RGB images. The dense depth maps are produced by accumulating LiDAR measurements from the whole sequences. The whole dataset is officially split to 86000 for training and 76000 for testing.
	
	\textbf{Make3D.}
	Make3D \cite{saxena_make3d_2009} dataset is hybrid scenes contains 1000 outdoor scenes and 50 indoor scenes. The whole dataset consists of 400 training images and 134 testing images with the resolution of $ 2272 \times 1704 $ pixels. The corresponding ground-truth depth maps are gathered by a custom 3D scanner with the image resolution of only $ 305 \times 55 $ pixels. 
	
	\subsection{Remote sensing benchmarks}
	
	\textbf{ISPRS dataset.}
	The ISPRS benchmark is originally designed for 2D Semantic Labeling contest \footnote{\url{https://www2.isprs.org/commissions/comm2/wg4/benchmark/semantic-labeling/}}. However, the DSM data are also provided in the Vaihingen and Postsdam datasets, which can thus also be utilized as benchmarks for height estimation.
	
	The Vaihingen dataset consists of 33 image tiles.
	The spatial resolution is 9 $cm$. Three bands are provided, i.e. near infrared (NIR), red (R), and green (G).
	The corresponding DSM is generated by dense image matching with a 9 $cm$ ground sampling distance using Trimble INPHO software.
	There are 16 image tiles for training and the remaining 17 images are used for testing.
	
	The Potsdam dataset is composed of 38 image tiles with spatial resolution of 5 $cm$.
	The aerial images consists of four bands, i.e. near infrared (NIR), red (R), green (G), and blue (B).
	The corresponding DSM is also generated by dense image matching using Trimble INPHO software, with ground sampling distance of 5 $cm$.
	24 tiles are used for training and the rest are for testing.
	
	\textbf{GRSS dataset.}
	The GRSS dataset is originally used in the 2018 GRSS Data Fusion Contest (DFC 2018) \footnote{\url{http://dase.grss-ieee.org/}}. The dataset contains a collection of multi-source data over Houston, US. It includes very high resolution color images resampled at 5cm/pixel, hyperspectral images, and DSM and DEM data derived from LiDAR data.
	There are 4 tiles available for training and the rest 10 are remained undisclosed for evaluation online.
	
 ReDWeb~\cite{xian_monocular_2018}.
	\subsection{Evaluation metrics}
	
	In order to evaluate and compare the performance of various depth estimation networks, a commonly accepted evaluation metrics are  categorized into errors ( rmse, rel, sqrel)  and accuracy. Errors are the lower, the better. The accuracy is the higher, the better. These indicators are formulated as:
	
	The root mean square error (rmse):
	\begin{equation}
	rmse = \sqrt{\frac{1}{N} \sum_{i=1}^{N}\left(\hat{d}_{i}-d_{i}\right)^{2}}.
	\end{equation}
	
	The mean relative error (rel):
	\begin{equation}
	rel = \frac{1}{N} \sum_{i=1}^{N} \frac{\left\|\hat{d}_{i}-d_{i}\right\|}{d_{i}}.
	\end{equation}
	The squared relative error (srel):
	
	\begin{equation} sqrel=\frac{1}{N} \sum_{i=1}^{N} \frac{{\left\|\hat{d}_{i}-d_{i}\right\|}^{2}}{d_{i}}\end{equation}
	
	%
	
	The accuracy is represented with the percentage of the relative depth prediction within threshold $1.25^{j}$:
	\begin{equation}
	\delta_{j}=\frac{\operatorname{N}\left(\hat{d}_{i} : \max \left\{\frac{\hat{d}_{i}}{d_{i}}, \frac{d_{i}}{\hat{d}_{i}}\right\}<1.25^{j}\right)}{\operatorname{N}\left(d_{i}\right)},
	\end{equation}
	where $d_{i}$ and $\hat{d}_{i}$  are  the ground truth depth values and the predicted ones respectively, and $N$ is the number of element of a set, $j={1,2,3}$. A higher $\delta_{i}$ indicates better prediction.
	For depth completion tasks on KITTI dataset, the officially evaluation metrics are : root mean square error ($RMSE$), mean absolute error ($MAE$),  root mean squared error of the inverse depth ($iRMSE$), and mean absolute error of the inverse depth ($iMAE$). They are formulated as below:
	\begin{equation}
	RMSE (\operatorname{mm})= \sqrt{\frac{1}{|\mathcal{N}|} \sum_{i \in \mathcal{N}}\left|\hat{d}_{i}-d_{i}\right|^{2}}
	\end{equation}
	
	\begin{equation}
	MAE(\mathrm{mm})= \frac{1}{|\mathcal{N}|} \sum_{i \in \mathcal{N}}\left|\hat{d}_{i}-d_{i}\right| 
	\end{equation}
	
	\begin{equation}
	iRMSE (1 / \mathrm{km})=\sqrt{\frac{1}{|\mathcal{N}|} \sum_{i \in \mathcal{N}}\left|1 / \hat{d}_{i}-1 / -d_{i}\right|} 
	\end{equation}
	
	\begin{equation}
	iMAE (1 / \mathrm{km})= \frac{1}{|\mathcal{N}|} \sum_{i \in \mathcal{N}}\left|1 / \hat{d}_{i}-d_{i}\right|
	\end{equation}

	\section{Supervised learning-based methods} \label{sec:sup}
	
	In this section, we introduce the supervised learning based monocular depth prediction methods from two perspectives: + regression-based methods and classification-based methods. The performance of the supervised learning-based methods are shown in Table \ref{tab:NYU} and Table \ref{tab:KITTI}.
	
	\begin{table}[]
		\centering
		\caption{The performance of supervised methods on NYU dataset.}
		\label{tab:NYU}
		\resizebox{0.8\textwidth}{!}{%
			\begin{tabular}{@{}lccccccccc@{}}
				\toprule
				Methods                                        & Year & Strategy       & rmse $\downarrow$   & rel$\downarrow$      & 1.25$\uparrow$  & $1.25^{2}$$\uparrow$ & $1.25^{3}$$\uparrow$ \\ \midrule
				Eigen et al. \cite{eigen2014depth}              & 2014 & regression     & 0.907  & 0.215   & 0.611 & 0.887      & 0.971      \\
				Zhuo et al. \cite{zhuo2015indoor}               & 2015 & regression     & 1.04   & 0.305   & 0.525 & 0.838      & 0.962      \\
				Wang et al. \cite{wang2015towards}              & 2015 & regression     & 0.824  & 0.22    & 0.605 & 0.89       & 0.97       \\
				Li et al. \cite{li2015depth}                    & 2015 & regression     & 0.821  & 0.232   & 0.621 & 0.886      & 0.968      \\
				Liu et al. \cite{liu2015deep}                   & 2015 & regression     & 0.759  & 0.213   & 0.65  & 0.906      & 0.976      \\
				Eigen et al. \cite{eigen2015predicting}         & 2015 & regression     & 0.641  & 0.158   & 0.769 & 0.95       & 0.988      \\
				Laina et al. \cite{laina2016deeper}             & 2016 & regression     & 0.79   & 0.194   & 0.629 & 0.889      & 0.971      \\
				Kim et al. \cite{kim2016unified}                & 2016 & regression     & 0.711  & 0.201   & 0.69  & 0.91       & 0.979      \\
				Dharmasiri et al. \cite{dharmasiri2017joint}    & 2017 & regression     & 0.624  & 0.156   & 0.776 & 0.953      & 0.989      \\
				Xu et al. \cite{xu2017multi-scale}              & 2017 & regression     & 0.583  & 0.121   & 0.811 & 0.954      & 0.987      \\
				Fu et al.\cite{fu2017a}                         & 2017 & regression     & 0.564  & 0.136   & 0.807 & 0.957      & 0.992      \\
				Li et al. \cite{li2017single}                   & 2017 & classification & 0.519  & 0.125   & 0.835 & 0.967      & 0.991      \\
				Yan et al. \cite{yan2017monocular}              & 2017 & regression     & 0.502  & 0.135   & 0.813 & 0.965      & 0.993      \\
				Cao et al. \cite{8010878}                       & 2018 & classification & 0.747  & 0.211   & 0.706 & 0.92       & 0.97       \\
				Alhashim et al. \cite{alhashim2018high}         & 2018 & regression     & 0.65   & 0.123   & 0.846 & 0.974      & 0.994      \\
				Xu et al. \cite{xu2018structured}               & 2018 & regression     & 0.593  & 0.125   & 0.806 & 0.952      & 0.986      \\
				Xu et al. \cite{8578175}                        & 2018 & regression     & 0.582  & 0.12    & 0.817 & 0.954      & 0.987      \\
				Lee et al. \cite{lee2018single-image}           & 2018 & regression     & 0.572  & 0.139   & 0.815 & 0.963      & 0.991      \\
				Heo et al.\cite{heo2018monocular}               & 2018 & regression     & 0.571  & 0.135   & 0.816 & 0.964      & 0.992      \\
				Qi et al. \cite{qi2018geonet:}                  & 2018 & regression     & 0.569  & 0.18    & 0.834 & 0.96       & 0.99       \\
				Nekrasov et al. \cite{nekrasov2018real-time}    & 2018 & regression     & 0.565  & 0.149   & 0.79  & 0.955      & 0.99       \\
				Zhang et al.\cite{zhang2018deep}                & 2018 & regression     & 0.54   & 0.134   & 0.83  & 0.964      & 0.992      \\
				Fu et al.\cite{fu2018deep}                      & 2018 & classification & 0.509  & 0.115   & 0.828 & 0.965      & 0.992      \\
				Zhang et al.\cite{zhang2018joint}               & 2018 & regression     & 0.501  & 0.144   & 0.815 & 0.962      & 0.992      \\
				Zhang et al. \cite{zhang2018progressive}        & 2018 & regression     & 0.501  & 0.144   & 0.835 & 0.962      & 0.992      \\
				Chen et al.\cite{chen2018rethinking}            & 2018 & regression     & 0.4871 & 0.114   & 0.852 & 0.971      & 0.997      \\
				Jiao et al. \cite{jiao2018look}                 & 2018 & regression     & 0.329  & 0.098   & 0.917 & 0.983      & 0.996      \\
				Hu et al.\cite{hu2019revisiting}                & 2019 & regression     & 0.635  & 0.143   & 0.788 & 0.958      & 0.991      \\
				Hu et al.\cite{hu2019revisiting}                & 2019 & regression     & 0.53   & 0.115   & 0.866 & 0.975      & 0.993      \\
				Chen et al.\cite{chen2019attention-based}       & 2019 & classification & 0.496  & 0.138   & 0.826 & 0.964      & 0.99       \\
				Yin et al.\cite{yin2019enforcing}               & 2019 & regression     & 0.416  & 0.108   & 0.875 & 0.976      & 0.994      \\ \midrule
				Cheng et al. \cite{ChengdepthVA}                & 2018 & sparse         & 0.117  & 0.016   & 0.992 & 0.999      & 1          \\
				Qiu et al. \cite{qiu2019deeplidar}              & 2019 & sparse         & 0.115  & 0.022   & 0.993 & 0.999      & 1          \\
				Xu et al.\cite{xu2019depthcom}                  & 2019 & sparse         & 0.112  & 0.018   & 0.995 & 0.999      & 1          \\
				Ma et al. \cite{ma2019self}                     & 2019 & sparse         & 0.23   & 0.044   & 0.971 & 0.994      & 0.998      \\
				Eldesokey et al. \cite{eldesokey2020confidence} & 2019 & sparse         & 0.123  & 0.017   & 0.991 & 0.998      & 1          \\
				Imran et al. \cite{imran2019depth}              & 2019 & sparse         & 0.131  & 0.013   & 0.868 & 0.954      & 0.979      \\
				Shivakumar et al. \cite{shivakumar2019dfusenet} & 2019 & sparse         & 0.176  & 0.037   & 0.987 & 0.998      & 1          \\
				Zhao et al. \cite{zhao2020adaptive}             & 2020 & sparse         & 0.105  & 0.015   & 0.994 & 0.999      & 1          \\
				Tang et al.\cite{tang2019learning}              & 2020 & sparse         & 0.101  & 0.015   & 0.995 & 0.999      & 1          \\ \bottomrule
			\end{tabular}%
		}
	\end{table}

	\begin{figure}[!ht]
		\centering
		\includegraphics[width=\linewidth]{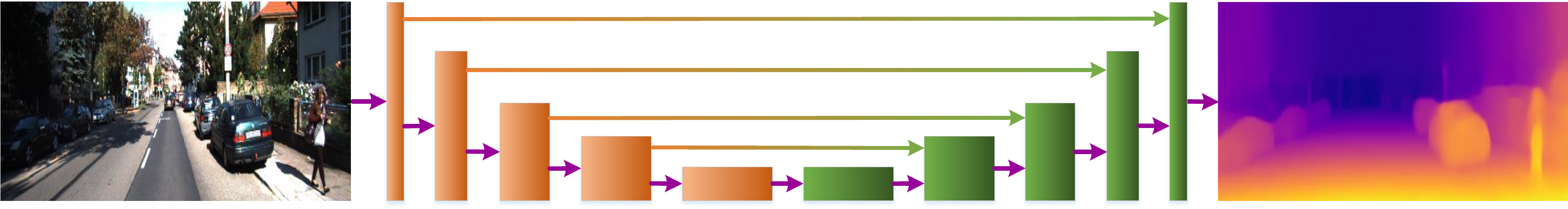}
		\caption{The   general network architecture of  supervised-based method for monocular depth prediction.}
		\label{fig:sup_network}
	\end{figure}
	
	\subsection{Regression-based methods}
	Regression-based methods directly model the hidden mapping function between RGB images and corresponding maps. We first give an introduction to the problem definition and then review the related works from single task learning , multi-task learning and learning with prior.
	
	\subsubsection{Single-task learning}
	\textbf{Problem formulation}
	Supervised learning-based monocular image depth prediction from RGB images can be regarded as regression problem. Let $I$ be the RGB image and $D$ is the depth map. Given a training dataset $T={(I_{i},D_{i})}^{M}_{i=1}$, $I_{i}\in I $ and $D_{i} \in D$, a CNN is trained to learn the hidden mapping function $\phi: I \rightarrow D$. The basic network architecture is comprised of an encoder and a decoder. The encoder consists of sequential convolutional blocks to extract the abstract information and the decoder contains several up-sampling modules to scale up the depth map resolution. A simple example is shown in Figure \ref{fig:sup_network}. The objective function is designed based on the deviation between the predicted depth values and the real ones. The most obvious manner is the normal distance like L1 norms or L2 norms distance. which are shown as below:
	
	\begin{equation}
	L_{1} = \frac{1}{N} \sum_{i=1}^{N}\left(\hat{d}_{i}-d_{i}\right).
	\end{equation}
	
	\begin{equation}
	L_{2} = \frac{1}{N} \sum_{i=1}^{N}{\left(\hat{d}_{i}-d_{i}\right)}^{2}.
	\end{equation}
	
	\textbf{Architecture and frameworks}
	Regression-based methods can be categorized into patch-based approaches and image-based approaches. Patch-based methods firstly separate the image into many patches or superpixels. Each superpixel is fed into  a CNN to predict a depth value. In this manner, each superpixel is associated with a depth label. Then some graphical models like CRF are used to produce pixel-level depth values. Li et al. \cite{li2015depth} obtain the super-pixels from an image and perform super-pixel level depth prediction regression using CNN. They refine the result to the pixel level with hierarchical conditional random filed. Two convolutional neural networks are leveraged to perform global depth inference and local depth inference in \cite{wang2015towards}. The results of two CNNS are fused through hierarchical CRF to get final pixel-level depth map. The global prediction allows to reduce the local ambiguities and the local prediction provides the detailed structure and boundaries. Both two methods utilize the CRF as the post-processing step.  In \cite{liu2015deep}, they also feed the image patches into the CNN. They take the output of the CNN as the unary and use the similarity between pixel as pairwise potentials, then design a CRF loss to train the network. Instead of using two-stage scheme, they combine the CNNs and CRF in a unified framework. Roy et al. \cite{roy2016monocular} predict depth map by combining the CNN with the random forest via regression. They slide over image with fixed window to generate image patch to represent the  pixels.  Every tree takes the image patch as input. The tree nodes are associated with a shadow CNN and the result of the CNN is passed to left child node or right one based on Bernoulli probability. The final depth value is estimated by fusion every leaf nodes with weights based on the corresponding path probabilities. Yan et al. \cite{yan2017monocular} also predict the depth from the superpixels and they use an auto-encoder to generate the pair-wise potential for CRF to refine the depth predictions.  
	
	Patch-based methods pay more attention to the local information while the global information are not well exploited. The global information is vital for the relative depth difference between pixels. Thus, depth estimation from the whole image becomes popular. Unlike the patch-based approaches, image-based methods take the whole image as input and predict the pixel-level depth directly. Eigen et al. \cite{eigen2014depth} firstly propose a coarse-to-fine frame with two convolutional neural network stacked together. The coarse-scale network consists of five convolutional neural blocks and two fully connected layers to extract global information. The fine-scale network refines the coarse depth prediction result with local information through a fully convolutional neural network. The fine-scale network takes the original image and coarse-scale network output as input. Fu et al. \cite{fu2017a} utilize a cascaded strategy to obtain depth of fine depth map. They use two branches: regression branch predicts a low solution continuous depth and classification branch for a high solution discrete depth map. The regression branch is used to refine the classification branch to obtain better details. Those methods follows a coarse-to-fine strategy to obtain fine-grained depth map. The refining stage is conducted either with a convolutional neural network or processing with CRF. Chakrabarti et al. \cite{chakrabarti2016depth} utilize a local to global optimization strategy to predict the depth map. They train a network to predict the depth derivatives distribution of different orders of each image location and globally optimize them to obtain the final depth map. Lee et al. \cite{lee2018single-image} utilize a two stream neural network to predict the depth and depth gradient, and then fuse them together in a convolutional block. The network is trained with depth deviation and gradient deviation between estimated depth maps and the ground truth. Zhang et al. \cite{zhang2018deep} propose a hierarchical depth guidance network and a hierarchical regularization learning method for fine-grained depth prediction. They utilize multi-scale supervision signal to extract depth representations and pixel-wise depth deviation, where gradient and depth value distribution consistency are used to regularize the network.
	
	New modules are designed to extract depth-related information from images. Laina et al. \cite{laina2016deeper} design a fully convolutional residual network for depth prediction. They use popular residual neural network to extract global information and design an efficient residual-upsampling block to obtain higher resolution depth map. Heo et al. \cite{heo2018monocular} design a new upsampling  filter called WSM to exploit the facts that depth vales are near in horizontal or vertical directions. The WSM filter adopts long rectangular kernels and replicates the kernel response in the horizontal or vertical direction. They also predict the confidential probability with the CNN. The final depth map is refined through CRF with predicted depth and confidential map. Chen et al. \cite{chen2020improving} exploit the structural information of image by introducing a novel attention module and a focal relative loss. The novel attention module is design to guide the network to focus on global structures  or local details. The focal relative loss penalizes the error depth discontinuous region by comparing the order of two points. 
	
	Different modules in CNNs is capable of extracting various representative information. Some researchers attempt to fuse them in multi-scale scheme for deriving depth related information. Xu et al. \cite{xu2017multi-scale} combine the CNN and continuous CRFs for depth prediction by exploiting multi-scale deep feature fusion through a CRF modules. They implement the mean field updates through sequential deep models and allow the network to be trained in end-to-end manner. They show that the depth prediction can benefit from the fusion of multi-scale deep features. They further introduce structure-attention-scheme to automatically determinate the information transferring between multi-scale features and allow the the network to train in end-to-end manner \cite{xu2018structured}. In  \cite{xu2018structured,xu2017multi-scale}, they either fuse the front-part  or end-part multi-scale information. In \cite{liu2020a}, they fuse front-part  multi-scale features with corresponding the end-part in the fashion of hourglass net. Instead of fusing them based on concatenation, they choose it with continuous CRF modules similar to \cite{xu2017multi-scale}. 
	
	Many researchers also refine the depth prediction with high order information like depth gradient or depth normal in direct or indirect manner. Wang et al. \cite{wang2016surge:} predict depth map, normal map, planar region and planar boundary with four CNN networks and the outputs are jointly constrained through a dense CRF module. Hua et al. \cite{hua2016depth} utilize two CNN networks to learn the features for absolute depth prediction and depth relations between individual pixel and its neighbours respectively. Then two features are fused in the conditional random field network to further refine the absolute depth prediction. Gan et al. \cite{gan2018monocular} explicitly model the relationships of different image locations with an asininity layer and combine absolute and relative features in an end-to-end network to exploit the relative features between neighbouring pixels. Besides, they introduce vertical pooling to aggregate image features vertically to improve the depth accuracy. Furthermore, they  enhance the depth labels by generating high-quality dense depth maps with off-the-shelf stereo matching method taking left-right image pairs as input. 
	
	Spatio-temporal consistency constraints enforce the  networks to ensure the predicted depth consistency of adjacent images. Kumar et al. \cite{kumar2018depthnet} exploit the spatio-temporal consistency between predicted depth map from an image sequence. They design a convolutional-LSTM-based network structure, which is trained by the depth difference in log space. Zhang et al. \cite{zhang2019exploiting} exploit the temporal depth consistency between consecutive frames of the video and propose a convolutional long-short memory network for it. They design a depth consistency loss based on adversarial learning that taking a depth sequence as input.  
	
	Pretrained on other dataset can also enlarge the generality of the netwoek Guo et al. \cite{guo2018learning} develop a method based on knowledge distillation. They use a stereo network to generate the proxy label for the monocular depth prediction. To avoid tremendous labelling work, they pre-train the stereo network with synthetic dataset and fine-tune it with the real one. Then the output disparity map acts as the supervision signal for monocular depth prediction. Alhashim et al. \cite{alhashim2018high} exploit transfer learning strategy by utilizing the pre-trained model on ImageNet as encoder for feature extraction. 
	
	To improve the computational efficiency, Wofk et al. \cite{wofk2019fastdepth:} build a lightweight network for depth prediction on embedding systems. They use MobileNet as encoder and devise a decoder that is comprised of five upsampling layers. Each layer contains a convolutional operation with nearest-neighbor interpolation and depth-wise decomposition. Ramamonjisoa et al. \cite{ramamonjisoa2020predicting} improve localization of the occlusion boundaries  of current monocular depth prediction approaches by re-sampling smooth depth map  with the displacement maps.
	
	\textbf{Loss}
	Supervised learning-based monocular image depth prediction trains  a convolutional neural network with various objective functions designed based on the deviation between the predicted depth values and the real ones. The most obvious manner is the normal distance like L1 or L2 norms distance.  To benefit from both losses, reverse Huber is used in \cite{laina2016deeper}. BerHu loss is equal to L1 loss when the residual is smaller than the given threshold $c$ and is L2 loss otherwise. Eigen et al. \cite{eigen2014depth,eigen2015predicting} design the loss function based on the distance between predicted depth and real ones in log space, which can help address the scale-problem. To enlarge the influence of the distant regions, Jiao et al. \cite{jiao2018look} put more weights on distant regions to reduce distribution bias by setting a coefficient that is linearly correlated to the ground truth depth values. Yin et al. \cite{yin2019enforcing} exploit normal constraints in 3D space to regularize the network for depth prediction. Instead of using a network to predict the normal, they compute it by rendering the depth map into point cloud based on the pinhole model. To preserve the border between different objects, Zhang et al. \cite{zhang2018progressive} devise a convolutional neural network trained with hard-mining loss. The hard-mining loss pays more attention on the hard region that easily accumulated errors while the rest is the same as berhu loss. Lee et al. \cite{lee2018single-image} propose a depth balance loss to prevent the network from deviation from near objects by using the two order polynomial. They generate multiple depth map candidates by cropping the input images and fusing them in Fourier domain to obtain the final depth map. Kuznietsov et al. \cite{kuznietsov2017semi} jointly utilize image reconstruction loss under Gaussian smoothing and depth deviation to train the network. To alleviate the blurry reconstruct of objects, Hu et al. \cite{hu2019revisiting} propose a new module to fuse multi-scale information of encoder using up-projection, channel-wise concatenation, a refine module to fuse information from decoder, and multi-scale fusion modules.  They combine the  difference in gradients and surface normal with balance Euclidean loss in log space to train the networks. Liang et al. \cite{liang2018learning} predict the disparity map of stereo images in feature space based on the deep feature constancy with depth-wise loss. Amiri et al. \cite{amiri2019semi} utilize both supervised loss and unsupervised loss as well as left-right consistency for depth prediction.
	
	Chen et al. \cite{chen2016single} design a depth order loss to replace the depth deviation for depth prediction. It can be represented as below: 
	\begin{equation}L(I, R, z)=\sum_{k=1}^{K} \psi_{k}\left(I, i_{k}, j_{k}, r, z\right)\end{equation}
	where $z$ is the predicted depth map, $R={(i_{k},j_{k},r_{k})}$, $k=1,...,K$  and $z_{i_{k}}$, $z_{j_{k}}$ are the depth of the location $i_{k}, j_{k}$, $r_{k} \in\{+1,-1,0\}$ is the ground truth depth order of $i_{k}$ and $j_{k}$: closer(+1), further (-1), and equal (0), $\psi_{k}\left(I, i_{k}, j_{k}, r, z\right)$ is the loss of the k-th query, 
	\begin{equation}\psi_{k}\left(I, i_{k}, j_{k}, z\right)=\left\{\begin{array}{ll}
	\log \left(1+\exp \left(-z_{i_{k}}+z_{j_{k}}\right)\right), & r_{k}=+1 \\
	\log \left(1+\exp \left(z_{i_{k}}-z_{j_{k}}\right)\right), & r_{k}=-1 \\
	\left(z_{i_{k}}-z_{j_{k}}\right)^{2}, & r_{k}=0.
	\end{array}\right.\end{equation}
	It is a ranking loss which encourages a small difference between depths if the $r_{k}=0$, otherwise encourages a large difference.
	
	Pixel-wise losses are considered conditionally independent from all other pixels and are generally local. They tend to produce blurry results since the losses are averaged in all pixels. However, the adversarial loss penalizes the joint configuration of all pixels and can be interpreted as a non-local loss, which can preserve more details. Chen et al. \cite{liu2019neural} utilize adversarial learning for the tasks to constraint the network with global optimization. Specially, they use conditional GANs, in which generator is used to produce the dense depth map and discriminator distinguish the RGB and depth patch pair between the real ones and generated ones. Kumar et al. \cite{kumar2018monocular} propose to utilize the adversarial learning to conduct depth prediction with adjacent images. They synthesize the neighbouring images with the predicted depth map and relative pose. The adversarial loss is jointly assisted with the photometric loss to train the model. Instead of directly designing loss based on the depth values, Liu et al. \cite{liu2019neural} predict the distribution of depth value and their confidence. They iteratively update the depth value and uncertainty over video clip in sequence. The results can be used for 3D reconstruction.

	\subsubsection{Multi-task learning}
	Apart from designing new modules to extract information from RGB images, learning related tasks in parallel could improve all the tasks over the each task alone. The rationale behind it is that multiple tasks are inherent related and they can constrain the network mutually for each task. Multi-task learning benefits the tasks by sharing certain network modules.  Eigen et al. \cite{eigen2015predicting} design a multi-scale network  trained for  depth, surface normal and semantic labels. Depth and surface normal are trained jointly by sharing the first scale stacks but differently with second and third scale stack. The semantic labels are trained with the same network architecture but with different weights. Semantic labels task is a higher order quantity compared to the depth and normals. The geometric related tasks are more related.  
	
	Dharmasiri et al. \cite{dharmasiri2017joint} develop a network jointly to estimate depth, surface normals and surface curves in the low level related tasks in parallel. With low-level hand crafted feature guiding the network, they achieve better performance than with supervision of two tasks alone. 	Qi et al. \cite{qi2018geonet:} incorporate geometric constraints between surface normal and depth to mutually benefit two tasks. On the one hand, surface normal is determined by local surface tangent plane of 3D points, which can be estimated from depth. On the other hand, depth is constrained by the local surface tangent plane determined by surface normal. They used two CNNs to predict initial depth and surface normal. Initial depth are refined given initial surface normal through kernel regression modules. The surface normal is updated through a residual module. With geometric constraints, they make the depth surface normal geometrically consistent and more accurate. Wang et al. \cite{wang2016surge:} also predict depth map, surface normal map separately with two convolutional neural networks. Further more, they estimate planar region and planar boundary with  CNN networks. The hidden geometric relation between  the outputs are jointly modelled through a dense CRF module for final depth and surface normal prediction. The method outperforms the CNN-based approach for depth and surface normal prediction alone.
	
	Depth estimation and semantic segmentation are two strongly correlated and mutually beneficial tasks.  Semantic labels allows  the depth prediction to be aligned  to the real physical size, and  depth enhance the semantic estimation via increasing the capacity of the classifier to distinguish the ambiguities.
	Mousavian et al.  \cite{mousavian2016joint} utilize a deep neural network for depth prediction and semantic segmentation simultaneously. The two tasks share the feature extractor and but of different decoders. A fully connected CRF layer is used to exploit the contextual relations of the two tasks. The performance of the two tasks are improved.
	Wang et al. \cite{wang2015towards} jointly predict the depth and semantic map in a neural network.  Region level depth prediction and semantic segmentation  are performed to be fused with global prediction through a two-layer Hierarchical Conditional Random Field. Accurate depth and semantic map are obtained  with clear edge and boundaries. Sanchez-Escobedo et al. \cite{sanchezescobedo2018hybridnet} also jointly perform semantic and depth prediction in a neural network. Instead of sharing the feature extractor, they use separate features extractors but mutually fuse the feature information in the extractor. They claim that such network architecture allows the task benefit from specific feature extraction for specific tasks and common feature for two tasks. Xu et al. \cite{8578175} perform multi-task learning through CNN to jointly predict depth estimation and Scene parsing. 	Instead of employing a cross modality interaction or designing effective joint-optimization objective function, they predict intermediate tasks like depth, surface normal, semantic and contours and use them for final depth prediction  and scene parsing. Kendall et al. \cite{kendall2017end-to-end} incorporate contextual information using 3-D convolutions over this volume. Further more,  a soft argmin function is proposed to train the network in an end-to-end fashion. Wang et al.   \cite{9157827} design a semantic divide-and-conquer framework for depth prediction. They categorize the scenes into semantic segments and predict  a normalized depth map for each semantic category separately. Then the final depth map is obtained by assembling the normalized depth maps by estimating  the scale and shift in global context.  	Zhang et al.  \cite{zhang2018joint} design a joint Task-Recursive Learning (TRL) framework for  semantic segmentation and monocular depth estimation tasks. TRL recursively refine the results of two tasks through serialized interactions. The interaction is fulfilled by designing  a specific Task-Attentional Module (TAM). Besides, TRL also incorporates the previous learning information for next network evolution to achieve fine-scale depth map progressively. Nekrasov et al. \cite{nekrasov2018real-time} propose a real-time network architecture to jointly predict semantic segmentation and depth estimation under uneven annotations conditions. They use a chained residual pooling module in a skipped connection manner to fuse encoder information for decoder. Besides, they use a knowledge distillation approach by introducing a teacher  from a trained model to address the asymmetric annotation problem. Cipolla et al. \cite{cipolla2018multi-task} simultaneously predict the semantic and instance semantic and depth in a single network . They develop an approach to automatically balance the weights of different learning objectives based on homoscedastic uncertainty. To analyse the cross modality influence between depth and semantic segmentation, they investigate modality influence numbers for performance in the depth and semantic segmentation. They find that large cross modality does not mean better accuracy and a beneficial balance is needed. They achieve good results on two tasks by optimizing the optimal hyper-parameter of the two tasks. 
	
	Besides, other related tasks are predicted jointly with CNNs. Ummenhofer et al.  \cite{ummenhofer2017demon:} utilize two image for depth prediction along with ego-motion by alternating optical flow estimation. Furthermore, they also design a scale invariant gradient loss with five different scales. Kim et al. \cite{kim2016unified} perform depth prediction and intrinsic image decomposition through new designed CNN network. The two tasks share the activations and layer, and refined with a gradient scale network.  Mancini et al.  \cite{8276580} design a specific structure for obstacles detection and depth prediction. The depth prediction branch is informed with object structures, which results in more robust estimations. On the other hand, the obstacle detection model exploits the depth information to predict obstacle distance and bounding boxes more precisely.

	\subsubsection{Learning with priors}
	Depth related information like camera parameters can help enhance the performance of CNNs-based depth prediction. He et al.   \cite{8360460} embed the focal length into the network design and fuse with deep feature information. Furthermore, they also generate a dataset of varying focal length from fixed focal length dataset for train the network. Facil et al.   \cite{facil2019cam-convs:} introduce camera intrinsic parameters to enable the network to learn the camera-independent pattern through a new convolutional module for supervised monocular depth prediction. The new convolutional module is comprised of the camera intrinsic parameters and centred coordinates channels. It enables the network to generalize to images captured with different cameras. Fei et al.  \cite{fei2019geo-supervised} exploit gravity prior to regularize the depth prediction. Given inertial information, the normal of the plane object can be computed. They design a loss based on the normal difference of plane object to penalize the network. 
	
	Synthetic datasets have been exploited for addressing depth labelling burden. Atapour-Abarghouei et al. \cite{atapourabarghouei2018real-time} propose to utilize the synthetic RGB-depth pairs based on transfer learning technique through adversarial learning framework  . Two models are trained for depth prediction and style transfer respectively. The depth model is trained with synthesizing depth and RGB image and the other one is trained to reduce the discrepancy between real data domain and synthesizing data domain.  To alleviate the depth labelling burden, Zhao et al.  \cite{zhao2019geometry-aware} exploit the synthetic images and epipolar geometry for depth estimation. They train two image style translators and depth estimator jointly based on domain adaption technology. Zheng et al. \cite{zheng2018t2net:} also propose to utilize the synthetic data to help depth prediction. Unlike previous methods using two networks to accomplish depth prediction and image style transfer, they jointly conduct the two tasks in one network. Kundu et al. \cite{kundu2018adadepth:} exploit synthetic datasets to solve the labelling tasks of supervised monocular depth prediction. To circumvent the domain gap, they use a unsupervised adversarial learning to mitigate it and regularize it with feature and domain consistency. Wu et al.  \cite{wu2019spatial} present a novel SC-GAN network with end-to-end adversarial training for depth estimation from monocular videos without estimating the camera pose and pose change over time. To exploit cross-frame relations, SC-GAN includes a spatial correspondence module which uses Smolyak sparse grids to efficiently match the features across adjacent frames, and an attention mechanism to learn the importance of features in different directions. Chen et al.  \cite{chen2018learning} propose a neural network to automatically assess the SFM reconstruction result for single view depth prediction. They generate a dataset called Youtube3D based on the network.

	\subsection{Classification-based methods}
	Many researchers cast the depth prediction as a classification problem. They group the depth values into many categories and train a CNN to determine which bins the pixels fall into. To obtain the continuous depth values, a post-processing step is conducted based on CRFs or probability scores. Cao et al.  \cite{8010878} discretize the  continuous depth values into several bins of the same size and use a CNN to determine bins pixels belonging to. Given the classification results and input image, a fully connected CRF is used to generate the final continuous depth prediction map. Li et al.  \cite{li2017single} divides the depth into bins in log space. Besides, they replace regular convolution with dilated convolution and a soft-weight-sum strategy is applied to convert the predicted score to continuous depth value. Chen et al. \cite{chen2019attention-based} also use a CNN to obtain depth classes and use the soft order inference to transform the predicted probability to continuous depth values . To further constrain the network, they also propose a attention loss to force the information consistency between RGB images and depth maps. Due to fact that minimizing mean squared error leads to slow convergence and local solution, Fu et al.  \cite{fu2018deep} propose to train the network with ordinary loss. They discretize the depth values equally in log space and cast the depth prediction as ordinal regression problem. They introduce a depth discretization strategy in log space to avoid over-strengthened loss on large depth values. Classification-based methods estimate the final dense depth map with a post processing step while the regression-based methods directly learns the depth values from the images with global and local information.

	Supervised learning-based methods is capable of restoring the scale information from single images and the structure of the scenes with high accuracy, given the fact that they directly train the network with the ground truth depth values. However, they require annotated datasets, which are expensive to obtain. To relieve the burden of labelling tasks, unsupervised learning is introduced.
	\section{Unsupervised learning-based methods} \label{sec:unsup}
	Unsupervised learning-based methods exploit the photometric constraints between monocular images sequence of stereo images instead of the deviation of the depth values.  The performance of unsupervised learning on KITTI dataset is shown in the bottom part of Table \ref{tab:KITTI}. In this section, we elaborate unsupervised learning-based methods from two aspects: reference the monocular video sequences and stereo image pairs.
	
	\begin{table}[]
		\centering
		\caption{The performance of supervised and unsupervised approaches on KITTI dataset.}
		\label{tab:KITTI}
		\resizebox{\textwidth}{!}{%
			\begin{tabular}{@{}lccccccccccc@{}}
				\toprule
				Methods                                       & Years & Strategy                  & rmse$\downarrow$   & rel$\downarrow$    & sqrel$\downarrow$   & 1.25$\uparrow$   & $1.25^{2}$$\uparrow$ & $1.25^{3}$$\uparrow$ \\ \midrule
				Eigen et al. \cite{eigen2014depth}             & 2014  & supervised/regression     & 7.156  & 0.19   & 1.515  & 0.692  & 0.899      & 0.967      \\
				Cao et al. \cite{8010878}                      & 2016  & supervised/classification & 6.311  & 0.18   & -      & 0.771  & 0.917      & 0.966      \\
				Kuznietsov et al. \cite{kuznietsov2017semi}    & 2017  & supervised/regression     & 4.751  & 0.117  & -      & 0.859  & 0.961      & 0.987      \\
				Fei et al.\cite{fei2019geo-supervised}         & 2018  & supervised/regression     & 5.734  & 0.143  & -      & 0.812  & 0.927      & 0.966      \\
				Kumar et al.\cite{xu2018structured}            & 2018  & supervised/regression     & 5.187  & 0.137  & 1.019  & 0.809  & 0.928      & 0.971      \\
				Xu et al.\cite{xu2018structured}               & 2018  & supervised/regression     & 4.677  & 0.122  & 0.897  & 0.818  & 0.954      & 0.985      \\
				Zhang et al.\cite{zhang2018deep}               & 2018  & supervised/regression     & 4.31   & 0.136  & -      & 0.833  & 0.957      & 0.987      \\
				Zhang et al.\cite{zhang2018progressive}        & 2018  & supervised/regression     & 4.082  & 0.136  & -      & 0.864  & 0.966      & 0.989      \\
				He et al.\cite{8360460}                        & 2018  & supervised/regression     & 4.014  & 0.086  & -      & 0.893  & 0.975      & 0.994      \\
				Gan et al. \cite{gan2018monocular}             & 2018  & supervised/regression     & 3.933  & 0.098  & 0.666  & 0.89   & 0.964      & 0.95       \\
				Guo et al.\cite{guo2018learning}               & 2018  & supervised/regression     & 3.258  & 0.09   & -      & 0.902  & 0.969      & 0.986      \\
				Chen et al.\cite{chen2018rethinking}           & 2018  & supervised/regression     & 2.349  & 0.061  & 0.282  & 0.943  & 0.988      & 0.996      \\
				Zhang et al.\cite{zhang2019exploiting}         & 2019  & supervised/regression     & 4.137  & 0.101  & -      & 0.89   & 0.97       & 0.989      \\
				Amiri et al.\cite{amiri2019semi}               & 2019  & supervised/regression     & 3.995  & 0.096  & 0.552  & 0.892  & 0.972      & 0.992      \\
				Chen et al. \cite{chen2019attention-based}     & 2019 & supervised/classification  & 3.599  & 0.083  & 0.437  & 0.919  & 0.982      & 0.995 \\
				Yin et al.\cite{yin2019enforcing}              & 2019  & supervised/regression     & 3.258  & 0.072  & -      & 0.938  & 0.99       & 0.998      \\
				Fu et al.\cite{fu2018deep}                     & 2019  & supervised/classification & 2.271  & 0.071  & 0.268  & 0.936  & 0.985      & 0.995      \\
				Guizilini et al.\cite{guizilini20193d}         & 2020  & supervised/regression     & 3.485  & 0.078  & 0.42   & 0.931  & 0.986      & 0.996      \\ \midrule
				Garg et al. \cite{garg2016unsupervised}        & 2016  & unsupervised/stereo       & 5.104  & 0.169  & -      & 0.74   & 0.904      & 0.958      \\
				Zhou et al.\cite{zhou2017unsupervised}         & 2017  & unsupervised/mono         & 5.181  & 0.201  & 1.391  & 0.696  & 0.9        & 0.964      \\
				Godard et al. \cite{godard2017unsupervised}    & 2017  & unsupervised/stereo       & 4.935  & 0.114  & -      & 0.861  & 0.949      & 0.976      \\
				Yang et al.\cite{yang2017unsupervised}         & 2018  & unsupervised/mono         & 6.641  & 0.1648 & 1.36   & 0.75   & 0.914      & 0.969      \\
				Pilzer et al.\cite{pilzer2018unsupervised}     & 2018  & unsupervised/stereo       & 6.187  & 0.166  & 1.466  & 0.757  & 0.906      & 0.961      \\
				Aleotti et al. \cite{aleotti2018generative}    & 2018  & unsupervised/stereo       & 5.998  & 0.119  & 1.239  & 0.846  & 0.94       & 0.976      \\
				Zhan et al.\cite{8578141}                      & 2018  & unsupervised/mono         & 5.583  & 0.151  & 1.257  & 0.81   & 0.936      & 0.974      \\
				Ranjan et al.\cite{ranjan2019competitive}      & 2018  & unsupervised/mono         & 5.326  & 0.14   & 1.07   & 0.826  & 0.941      & 0.975      \\
				Poggi et al.\cite{poggi2018learning}           & 2018  & unsupervised/stereo       & 5.205  & 0.126  & 0.961  & 0.835  & 0.941      & 0.974      \\
				Mahjourian et al.\cite{mahjourian2018unsupervised} & 2018 & unsupervised/mono      & 4.546  & 0.155  & 0.927  & 0.781  & 0.931      & 0.975 \\
				Yin et al.\cite{yin2018geonet:}                & 2018  & unsupervised/mono         & 4.348  & 0.147  & 0.936  & 0.81   & 0.941      & 0.977      \\
				Wong et al. \cite{wong2019bilateral}           & 2019  & unsupervised/stereo       & 5.515  & 0.133  & 1.126  & 0.826  & 0.934      & 0.969      \\
				Chen et al.\cite{chen2019towards}              & 2019  & unsupervised/stereo       & 5.096  & 0.118  & 0.905  & 0.839  & 0.945      & 0.977      \\
				Zhou et al.\cite{zhou2019unsupervised}         & 2019  & unsupervised/mono         & 4.945  & 0.121  & 0.837  & 0.853  & 0.955      & 0.982      \\
				Casser et al. \cite{casser2019unsupervised}    & 2019  & unsupervised/mono         & 4.7503 & 0.1087 & 0.825  & 0.8738 & 0.9577     & 0.9825     \\
				Chen et al. \cite{chen2019self-supervised}     & 2019  & unsupervised/mono         & 4.743  & 0.099  & 0.796  & 0.884  & 0.955      & 0.979      \\
				Tosi et al. \cite{tosi2019learning}            & 2019  & unsupervised/stereo       & 4.714  & 0.111  & 0.867  & 0.864  & 0.954      & 0.979      \\
				Watson et al. \cite{watson2019self-supervised} & 2019  & unsupervised/stereo       & 4.393  & 0.098  & 0.702  & 0.887  & 0.963      & 0.983      \\
				Puscas et al. \cite{puscas2019structured}      & 2019  & unsupervised/mono         & 4.223  & 0.1283 & 0.8681 & 0.84   & 0.941      & 0.971      \\
				Almalioglu et al. \cite{almalioglu2019ganvo:}  & 2019  & unsupervised/mono         & 3.671  & 0.137  & 0.892  & 0.808  & 0.939      & 0.975      \\
				Wang et al.\cite{wang2019unos:}                & 2019  & unsupervised/mono         & 3.404  & 0.049  & 0.515  & 0.965  & 0.984      & 0.992      \\
				Wang et al.\cite{wang2019recurrent}            & 2019  & unsupervised/mono         & 2.32   & 0.112  & 0.418  & 0.882  & 0.974      & 0.992      \\
				Zhao et al.\cite{9156629}                      & 2020  & unsupervised/mono         & 4.581  & 0.113  & 0.704  & 0.871  & 0.961      & 0.984      \\ \bottomrule
			\end{tabular}%
		}
	\end{table}

	\subsection{Stereo images-based methods}
\begin{figure}[]
		\centering
		
		\includegraphics[width=\linewidth]{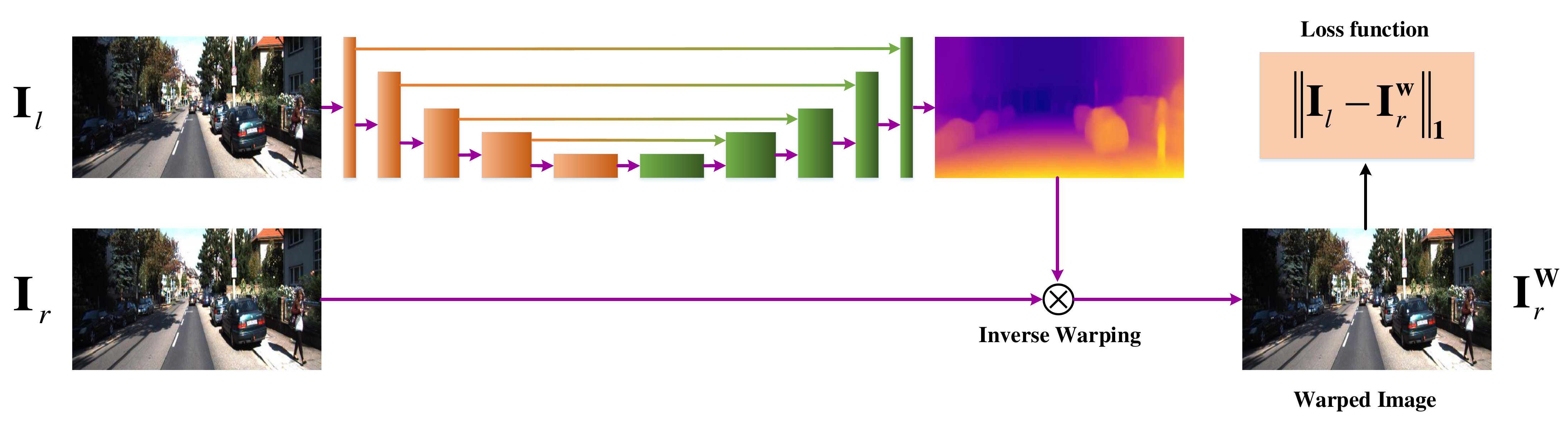}
		\caption{The   general network architecture of unsupervised-based method with stereo images.}
		\label{fig:stereoapproach}
		
	\end{figure}

	Stereo images-based methods train a convolutional neural network to estimate the depth map from the paired stereo images as shown in Figure \ref{fig:stereoapproach}. The predicted depth map is used to synthesize the left images or right images from the other ones through inverse warping. The difference between the synthesized images and the real ones are used to design the loss function for network training as shown in equation~\ref{photometricloss}. The inverse warping process can be expressed as:
	
	\begin{equation}\label{e:inversmapp}
	p_{l} \sim \mathbf{K} T_{r \rightarrow l} D_{r}\left(p_{r}\right) \mathbf{K}^{-1} p_{r}.
	\end{equation}
	
	where the $p_{r}$ represents the pixel on the right image and the $p_{l}$ indicates the corresponding pixel of the left image. $K$ is the camera intrinsic matrix, which is known. $D_{r}\left(p_{r}\right)$ stands for the depth value of the right image. $ T_{r \rightarrow l}$ refers to the transformation matrix from the right image to the left image. Given the depth map and the known transformation matrix, the pixels between left images and right images can be associated.  Based on such geometric projection relationship, we can synthesize images. Thus, the loss function of stereo images-based methods can be written as:
	
	\begin{equation}\label{photometricloss}
	\mathcal{L}_{v s}=\frac{1}{N} \sum_{L}^{N}\left|I_{n}(p)-\hat{I}_{n}(p)\right|.\end{equation}
	
	Various approaches have been designed for stereo-image based approach. Xie et al. \cite{xie2016deep3d:} train a convolutional neural network to synthesize a corresponding right view by taking the left view of stereo pairs as input. Supervised with photometric loss, they produce disparity distribution for pixels. The main disadvantage of the approach is high consumption of memory when scaling to larger output resolutions. Garg et al. \cite{garg2016unsupervised} utilize the photometric loss to estimate depth from stereo image pairs. They predict the dense depth map from one image and reconstruct other image with the input image given the predicted depth map and pre-calibrated stereo camera parameters.  They further include the depth gradient as the regularizer to obtain smooth depth map. However, their model is not fully differential.  Godard et al. \cite{godard2017unsupervised} extend depth image reconstruction loss by combining L1 distance and single scale SSIM in a fully differential manner. Furthermore, they also predict the depth map of the right-view image and constrain the network with left-right depth consistency loss. They also use a smoothness term to encourage locally smoothness with L1 penalty on depth gradient by weighting with edge information. Tosi et al. \cite{tosi2019learning} introduce the traditional stereo image depth prediction method to generate a proxy label as supervised signal with the manner of the reverse Huber loss along with image reconstruction loss and depth smoothness loss. They prove such assistance can enable the network to achieve better result. Wong et al. \cite{wong2019bilateral} propose a novel objective function account for the bilateral cyclic relationship between the left and right disparities and introduce an adaptive regularization scheme to handle both the co-visible and occluded regions in a stereo pair.
	
	However, reconstructing images with stereo images suffer from problems in occlusion and border regions. Poggi et al. \cite{poggi2018learning} address the problem by generating virtual trinocular view from stereo images. This is achieved by setting left and right view as center view respectively and trained with re-projection errors.   
	
	Since pixel-level objective function are easy locally minimum, Chen et al. \cite{chen2019towards} also exploit the semantic consistency for unsupervised depth prediction from stereo images. They design a unified decoder for both depth prediction and semantic segmentation. They design a semantic consistency loss to ensure the semantic consistency from left image to right image, and a semantic guided depth loss for the cross modality. Aleotti et al. \cite{aleotti2018generative} utilize a adversarial learning framework for depth prediction . They synthesize the warped target image with the inferred depth generated by the generator and trained with the adversarial loss. Pilzer et al. \cite{pilzer2018unsupervised} propose to use adversarial learning framework for unsupervised depth prediction . They design a cycled generative neural network to learn the depth and synthesize the image in a closed form, which enforces strong constraints for two generators.
	Zhai et al.  \cite{zhai2020an} utilize the object contextual information by designing a object context-aware network and jointly predict the depth and optical flow in a supervised manner. The object context-aware network feeds the predicted depth map and flow map as well as the second last feature maps into a series of dilated convolutional layers. 
	
	\subsection{Video sequence-based methods}
	\begin{figure}[!ht]
		\centering
		\includegraphics[width=\linewidth]{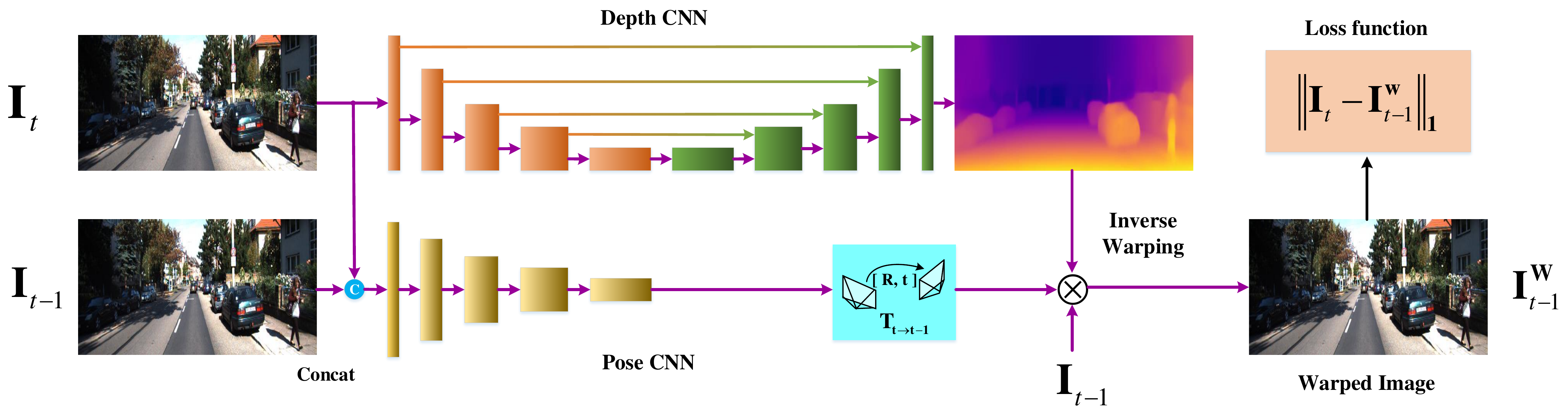}
		\caption{The   general network architecture of  unsupervised-based method with monocular videos.}
		\label{fig:monoapproach}
	\end{figure}
	Considering that collecting binocular stereo images requires special instruments, monocular videos can be a good alternative to stereo images.  Monocular videos-based methods also utilize the  photometric loss to supervise the training. Instead of taking stereo image pairs as input, they take multiple consecutive images as input and predict the corresponding depth map of the target image and relative poses between target image and adjacent images. The target image is reconstructed with the predicted depth, relative poses and nearby images. Unlike stereo image pairs, for which the relative pose is known and fixed, a separate CNN network is needed to obtain relative pose of two adjacent images. A typical network architecture of video sequence-based methods is shown in Figure \ref{fig:monoapproach}.
	
	Zhou et al. and Vijayanarasimhan et al.  \cite{zhou2017unsupervised} \cite{vijayanarasimhan2017sfm-net:} attempt to use monocular videos for unsupervised depth prediction with photometric loss.  Besides, they both exploit the smoothness term and depth gradient regularizer as penalties to train the network. Furthermore, depth consistency loss of adjacent images are used in \cite{vijayanarasimhan2017sfm-net:}.  Mahjourian et al. \cite{mahjourian2018unsupervised} propose a loss in 3D space to train the network . The 3D loss is designed based on the consistency of point clouds rendering from the estimated depth of adjacent images. They generate the point cloud from the estimated images and align the adjacent point clouds using ICP algorithm and the loss is comprised of transition loss and registration errors. The structured similarity (SSIM) is also employed to perform network regularization in patch levels. Yin et al. \cite{yin2018geonet:} refine the depth smoothness regularizer with depth gradient weighted terms to keep sharp edges. Since the occlusion regions or non-Lambertian surface violate the geometric consistency assumption, they only calculate the geometric consistency loss in effective regions. The occlusion and non-Lambertian regions are identified as the place that forward/backward optical flows contradict seriously.
	
	Some researchers utilize generative-adversarial network to further constrain the depth prediction from global perspective. Almalioglu et al.  \cite{almalioglu2019ganvo:} propose a network to perform monocular visual odometry and depth estimation simultaneously using adversarial learning to  provide supervision signals with photometric loss.  The generator estimates the depth map from the target image. The pose regressor predict the relative pose. The two output are used to synthesize the target image from multiple views. The discriminator discriminates the reconstructed image and original image to form the adversarial objective function.  Puscas et al. \cite{puscas2019structured} propose a dual GAN model for monocular depth predict. The two GAN are coupled using CRF modules and are trained in an end-to-end manner. The proposed network predicts the multiple poses, one depth image and reconstruct the target RGB image with multiple input images. The objective function consists of a photometric loss and an adversarial loss. 
	
	Photometric loss is meaningful under three assumptions: 1)the scene is static and no moving objects exists; 2) occlusion regions do not appear between the target images and the source images; 3) the surface is Lambertian. Therefore, regions do not meet the conditions should be detected. Zhou et al. \cite{zhou2017unsupervised} utilize an auto-encoder structure to predict explainability mask that indicates the motion regions  with pose estimation by sharing the feature extraction part  . Vijayanarasimhan et al.  \cite{vijayanarasimhan2017sfm-net:} predict the object masks to obtain the motion regions. 
	Gordon et al. \cite{gordon2019depth} utilize a united network for relative pose, camera intrinsic parameters and depth prediction instead of using two networks. They detect occlusion region based on geometric constraints in a differential manner with predicted depth map. Intrinsic parameters and motion object detection are addressed with randomized layer normalization. Besides, they automatically calculate the effective masks based on geometry principle instead of direct prediction without considering moving objects. Yin et al. \cite{yin2018geonet:} also perform a multi-task learning strategy to jointly learn dense depth, optical flow and camera pose. They detect the motion or occluded regions with predicted optical flow with the ResflowNet. Besides, they further utilize the forward and backward depth consistency to regularize the network. Casser et al. \cite{casser2019unsupervised} exploit the structure and semantic information by modelling the objects motion in 3D space as well as ego-motion and depth estimation for unsupervised monocular depth and ego-motion predictions. Ranjan et al. \cite{ranjan2019competitive} design a competitive and collaborative scheme to predict depth, ego-motion, optical flow and static and moving objects segmentation simultaneously.They divide the tasks into static scene reconstruction including depth prediction, ego-motion and moving regions reconstruction. They competitively achieve each tasks and adjust the motion segmentation works to work collaboratively. Wang et al. \cite{wang2019unos:} design a unified network to predict stereo depth, optical flow and motion jointly. They also take consideration of moving objects by constructing the soft masks of potential moving objects or occlusion. Zhan et al. \cite{8578141} propose a network to fulfil the single view depth prediction and visual odometry simultaneously in an unsupervised way . Reconstruction error between stereo images and adjacent frames are utilized to train the network. Instead of utilizing the photometric error, they demonstrate that reconstruction error based on deep features gives better result. Shu et al. exploit the feature-metric loss to enhance the unsupervised depth estimation \cite{shu2020feature}. The feature is learned through single-view reconstruction and the feature-metric loss is defined as the photometric loss in the learned feature space.
	
	Current unsupervised learning-based methods still suffer from the scale ambiguity issues. To address the problem of scale ambiguity for unsupervised depth prediction, Bozorgtabar et al. \cite{9010303} leverage the virtual depth and RGB images pairs to train the depth prediction model. To reduce the discrepancy between real image and virtual RGB image, they design a deep feature alignment loss. Besides, the photometric loss is  utilized to predict the relative pose as well. Wang et al.  \cite{wang2018learning} exploit the direct visual odometry (DVO) approaches for depth prediction and eliminate the pose CNN in unsupervised depth prediction. Instead, they use DVO approach to calculate the pose and address the scale ambiguity problem with a depth normalization before calculating the loss. Zhao et al.  \cite{9156629} develop a novel system that explicitly disentangles scale from the network estimation. Instead of relying on PoseNet architecture, they directly solve fundamental matrix from dense optical flow correspondence for relative pose estimation and make use of a two-view triangulation module to get the scale of 3D structure.  To obtain high resolution depth map, Zhou et al. \cite{zhou2019unsupervised} design a dual neural networks including low resolution network and high resolution networks . The low resolution network are used to extract abstract information and high resolution network to generate high resolution depth map. A self-assemble module is devised to flow information from low resolution network to high resolution network. 
	
	Some researchers investigate to achieve real-time efficiency on mobile devices. Poggi et al.  \cite{poggi2018towards} devise a CNN network architecture to employ the unsupervised depth estimation in mobile device or CPU in real-time. They use twelve convolutional layer to derive six level pyramid feature, which is achieved by setting the stride of the first layer as two. Using the network, they are able to obtain two image per second. Peluso et al. \cite{8714893} implement a real-time unsupervised depth prediction network on an ARMv7-based platform . The network structure uses the PyNet and quantify the number for optimization and neural kernel convolution to save storage space and accelerate computation speed. In \cite{liu2020mininet}, they obtain multi-scale information by designing the encoder in a recurrent manner and depth-wise convolutions are leveraged in decoder module to further reduce the parameters. Zhu et al. \cite{zhu2019unsupervised} utilize the event camera information to perform the depth prediction and relative pose estimation. 
	
	Temporal coherence between adjacent video frames can facilitate the depth prediction. Zhang et al.  \cite{zhang2019exploiting} design a network structure building on convolutional long short-term memory (LSTM) to exploit spatial and temporal information between videos. They utilize a 3D convolutional neural network as the discriminator by taking as input video sequences to regularize the depth generator.
	Wang et al. \cite{wang2019recurrent} embed a conLSTM module into depth prediction network and pose regression network, which are trained with re-projection error or depth loss along with optical flow consistency. It enhances the depth prediction performance in both unsupervised and supervised manners . Zhang et al.  \cite{9157014} design adapter modules to adjust feature representation without changing pre-learned weights in online case. They also exploit temporal depth consistency in objective functions and regularization in optimization procedure. Johnston et al.  \cite{johnston2020self} introduce a self-attention mechanism and perform discrete disparity prediction along with the depth estimation . Self-attention allows the network to learn a general contextual information specially for the similar patches at non-contiguous regions of the image. Discrete disparity estimation provides more robust and sharper results and enable the depth  uncertainty estimation.
	
	Photometric loss fails to work in the textureless indoor scenes. To address it, Zhou et al.  \cite{zhou2019moving} propose to exploit the optical flow  to incorporate with the photometric loss to train the network. They use an additional module to generate the optical flow as the supervise signal to supervise the depth prediction process and pose regression process. 
	Pilzer et al.  \cite{pilzer2019refine} exploit cycle inconsistency and knowledge distillation for unsupervised depth prediction. The cycle-inconsistency of two adjacent images and the discrepancy between teacher network and student network are used to train the network. 
	
	Additional information has been exploited to enforce the network for accurately depth prediction. Watson et al.  \cite{watson2019self-supervised} use the depth generated from Semi-Global Matching algorithm of stereo pairs to assist neural network to get rid of the local minimum caused by the photometric loss. Yang et al.  \cite{yang2017unsupervised} leverage the compatibility between depth map and surface normal to train the network. They firstly use a network to obtain the coarse depth map and generate the normal map, then the normal map are used to refine the coarse map through a normal-to-depth layer. Prasad et al.  \cite{prasad2019sfmlearner} improve the photometric loss with epipolar constraints. They add the weights to each pixels according to fitness of epipolar geometry. Zhou et al.  \cite{zhou2018unsupervised} exploit the depth consistency constraints from different views to regularize the network. They also generate large depth map using super-resolution network. Motivated by the bundle adjustment, Chen et al. \cite{chen2019self-supervised} jointly perform depth, optical flow, relative pose and intrinsic parameters estimation in a convolutional neural network . Photometric and geometric constraints are exploited to regularize the network. Zou et al.  \cite{zou2018df} leverage geometric consistency as additional supervisory signals for the problem. The core idea is to synthesize 2D optical flow by back-projecting the induced 3D scene flow with the predicted scene depth and camera motion. The loss is constructed based on discrepancy between the rigid flow and the estimated flow, which allows to impose a cross-task consistency. Guizilini et al. \cite{guizilini20193d} leverage 3D convolutions to learn representations that maximally propagate dense appearance and geometric information. A novel loss is designed based on the camera’s velocity and allows to solve the inherent scale ambiguity in monocular vision .
	
	Unsupervised learning-based approaches significantly decrease the effort on the depth labelling, but it suffer from the scale ambiguity and low accuracy.  To further increase the accuracy, sparse depth samples are fused with the RGB images for depth estimation.

	\section{Sparse samples guided approaches} \label{sec:sparse}
	Sparse depth map guided approaches aim at generating a dense depth map from RGB image and a sparse depth map. We group the approaches into three classes according to the fusion strategies: early fusion and later fusion. The performance is shown in Table \ref{tab:sparseKitt} on KITTI dataset. Firstly we give the general idea of the problem.

	\subsection{Problem formulation}
\begin{figure}[!ht]
		\centering
		\includegraphics[width=\linewidth]{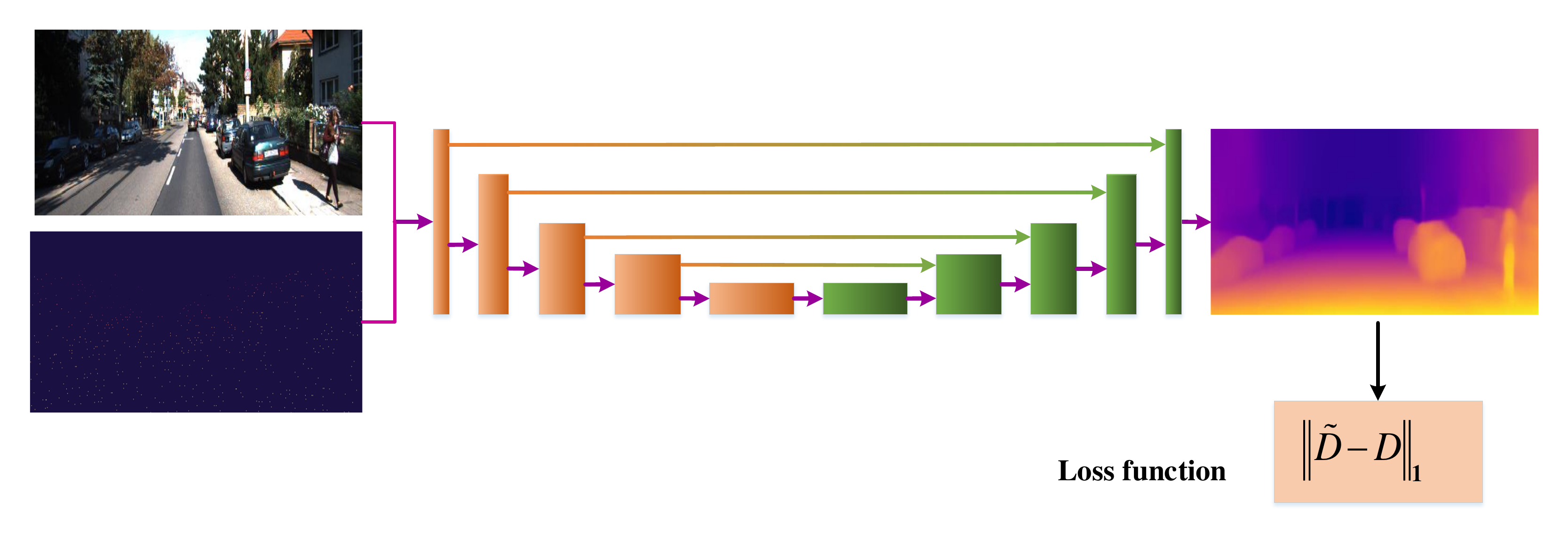}
		\caption{The   general network architecture of  monocular depth prediction with sparse sample guidance.}
		\label{fig:depthcomletion}
	\end{figure}

	Given an image $I$ and a sparse depth map $S$, the dense depth map $D$ is predicted from the depth estimation function $f$ by minimising certain loss function $L$.  $f$ is modelled by a convolutional neural network and parameterized by $\theta$. Figure \ref{fig:depthcomletion} gives a basic network structure for the approaches. The loss function is also formulated based on the deviation between the predicted depth map and the real ones when the ground-truth depth map is available. Otherwise, it can be trained in self-supervised manner where the $L$ is the combination of the L1 loss and the SSIM loss coupled with a smoothness regularizer. These approaches will be classified to two categories based on where RGB image and sparse depth map are fused.
	\begin{equation}\theta^{~}=\arg \min _{\theta} \mathcal{L}(f(I, S ; \theta), D_{g})\end{equation}
	
	\begin{table}[]
		\centering
		\caption{The performance of sparse samples guided methods on KITTI dataset.}
		\label{tab:sparseKitt}
		\resizebox{0.7\textwidth}{!}{%
			\begin{tabular}{@{}llllll@{}}
				\toprule
				Methods                             & Year & RMSE    & MAE    & iRMSE & iMAE \\ \midrule
				Cheng et al. \cite{ChengdepthVA}     & 2018 & 1019.64 & 279.46 & 2.93  & 1.15 \\
				Jaritz et al.\cite{jaritz2018sparse} & 2018 & 917.64  & 234.81 & 2.17  & 0.95 \\
				Dimitrievski et al. \cite{dimi2018learning}      & 2018 & 1045.45 & 310.49 & 3.84 & 1.57 \\
				Shivakumar et al. \cite{shivakumar2019dfusenet}  & 2019 & 1206.66 & 429.93 & 3.62  & 1.79 \\
				Imran et al. \cite{imran2019depth}   & 2019 & 965.87  & 215.75 & 2.43  & 0.98 \\
				Zhang et al.\cite{zhang2019dfinenet} & 2019 & 943.89  & 304.17 & 3.21  & 1.39 \\
				Yang et al.  \cite{yang2019dense}    & 2019 & 832.94  & 203.96 & 2.1   & 0.85 \\
				Eldesokey et al. \cite{eldesokey2020confidence}          & 2019 & 829.98  & 233.26 & 2.6  & 1.03 \\
				Ma et al. \cite{ma2019self}        & 2019 & 814.73  & 249.95 & 2.8   & 1.21 \\
				Xu et al. \cite{xu2019depthcom}          & 2019 & 777.05  & 235.17 & 2.42  & 1.13 \\
				Gansbeke et al.\cite{gansbeke2019sparse} & 2019 & 772.87  & 215.02 & 2.19  & 0.93 \\
				Qiu et al. \cite{qiu2019deeplidar}   & 2019 & 758.38  & 226.5  & 2.56  & 1.15 \\
				Chen et al.\cite{chen2019learning}   & 2019 & 752.88  & 221.19 & 2.34  & 1.14 \\
				Tang et al. \cite{tang2019learning}  & 2019 & 736.24  & 218.83 & 2.25  & 0.99 \\
				Wong et al.\cite{wong2020unsupervised}                   & 2020 & 1169.97 & 299.41 & 3.56 & 1.2  \\
				Lopez-Rodriguez et al. \cite{lopez-rodriguez2020project} & 2020 & 1062.48 & 268.37 & 3.12 & 1.13 \\
				Schuster et al. \cite{2020ssgp}      & 2020 & 838.22  & 244.7  & 2.51  & 1.09 \\
				Xu et al. \cite{xu2020deformable}    & 2020 & 766.74  & 220.36 & 2.47  & 1.03 \\
				Li et al.\cite{li2020a}              & 2020 & 762.19  & 220.41 & 2.3   & 0.98 \\
				Zhao et al. \cite{zhao2020adaptive}  & 2020 & 744.91  & 206.09 & 2.08  & 0.9  \\
				Cheng et al. \cite{ChengWGY20CSPNLC} & 2020 & 743.69  & 209.28 & 2.07  & 0.9  \\
				Park et al.\cite{park2020non}   & 2020 & 741.68  & 199.59 & 1.99  & 0.84 \\ \bottomrule
			\end{tabular}%
		}
	\end{table}
	
	\subsection{Early fusion}
	Early fusion strategy takes the concatenated RGBD data as input. The sparse depth map are firstly organized as the same resolution of the RGB images by encoding the invalid position as $0$ to address the irregular pattern of the sparse map. Then the RGB images and the formatted map are concatenated as a four channel input. Some researchers improve the performance by parameterizing the representation of the sparse input depth and output representation. Chen et al. \cite{chen2018estimating} develop a  flexible invertible approach to parameterize the sparse depth input, which achieves comparable results to the conventional depth sensors. Nearest neighbour filling and Euclid distance transform are used to produce the dense initial depth map and the sparse depth pattern masks, respectively. Nearest neighbour filling provides initial depth information at each point and Euclid distance transformation map includes depth pattern mask and residual magnitude information. The two parameterized the depth information are stacked with the RGB images as the input for the multi-scale neural network to generate the dense depth map. Imran et al.  \cite{imran2019depth} utilize a one-hot representation for depth values and train the networks with cross-entropy loss, which enable the network to address the inter-object depth mixing.
	
	Various architectures are designed to explore the mutual guidance of the two modalities. Qu et al.  \cite{qu2020depth} design a least square fitting modules to replace the $1\times 1$ convolutional layer to  fit the implicit depth bases to the sparse depth target. The module can be extended to multi-scale formulation to improve training. They were fused together with the RGB image in a CNN network. A new building block is designed to perform RGBD fusion in 2D and 3D space respectively in \cite{chen2019learning}. The 2D branch performs traditional convolution to extract appearance information and the 3D branch does the continuous convolution based on K-nearest neighbour. The features from 2D-3D branches are fused in hierarchical way to fully exploit the complementary information of the two representations. Ma et al.   \cite{ma2019self} propose a framework with two RGB images assistance with photometric reconstruction loss. One image is fused with sparse depth map to generate dense depth map while another one is used to reconstruct the first image with predicted depth and relative pose computed from PnP algorithm. Fu et al. \cite{fu2019lidar} utilize the skip connection and design a new residual Up-projection module to recover the depth prediction to targeted resolutions .  Dimitrievski et al.  \cite{dimi2018learning} design a morphological  neural network to complete the sparse depth map with the RGB guidance using early fusion U-net structure. They approximate the morphological operation with Contra-Harmonic Mean filter. The resulting depth images preserve clear object boundaries. Another direction is to learn affinity from RGB for spatial diffusion for the sparse depth map. Park et al.  \cite{park2020non} design a non-local spatial propagation network to simultaneously estimate the initial depth map, pixel-wise uncertainty and the non-local neighbours associated with affinities. The initial depth map is iteratively refined with predicted uncertainty map and non-local neighbours. Non-local neighbours are iteratively updated to avoid irrelevant ones.  Chen et al. \cite{ChengdepthVA} design a convolutional spatial propagation network (CSPN) to learn the affinity kernel weights for depth completion . The kernel is operated on the neighbours pixels recurrently. They further refine it by learning the kernel size and number of iteration in \cite{ChengWGY20CSPNLC}.  Xu et al.\cite{xu2020deformable} directly takes the RGB images and Sparse LiDAR maps as input and estimate the coarse depth map and confidence mask . Furthermore, they design a deformable propagation network to refine the networks. They use the convolutional neural network to adaptively determine the receptive field of each pixel as well as the affinity kernel weights. They derive the more relevant pixels for propagation.
	
	\subsection{Late fusion}
	Late fusion strategy extracts features from the sparse depth map and RGB image, separately. Then, the extracted features are fused through CNN for final depth completion. Combining them late at the feature level is a better choice \cite{jaritz2018sparse}. Some methods extract specific features from an image to assist in dense estimation. Hua and Gong   \cite{hua2018a} design a normalized convolutional layer to fully address the sparse and irregular of the sparse points, and the RGB image are coped with the regular CNNs. The two information are fused together through standard CNN for dense depth map estimation. Gansbeke et al.  \cite{gansbeke2019sparse} use two CNNs to derive global and local information respectively from concatenated RGB and LiDAR map. The global information is approached as the prior to guide the local information extraction and lately fuse two source information for depth completion and noise correction. Jaritz et al.  \cite{jaritz2018sparse} handle sparse depth data with optional dense RGB, and accomplish depth completion and semantic segmentation changing only the last layer. Shivakumar et al. \cite{shivakumar2019dfusenet} design a two branches network structure to extract contextual information from sparse depth map and RGB image respectively and fuse together with a sequence of de-convolutions to recover the depth scale. 
	Qiu et al. \cite{qiu2019deeplidar} take the late fusion strategy and fuse the RGB and the sparse information only in the decoder . Instead of directly using the RGB images, they produce the surface normal as the intermediate representation. Besides, they predict the confidence mask for occlusion handling and attention maps to integrate the sparse depth, color image and normal maps. Eldesokey et al. \cite{eldesokey2020confidence} use two branches to extract information from sparse LiDAR data and RGB images separately. They feed the confidence map of the LiDAR data with the two module data for feature extraction. Then the output two branches are concatenated as input of fusion net to produce the dense map. The normalized convolution layer is employed to address the unstructured depth data. 
	Huang et al.  \cite{huang2020hms} extend the concept of sparsity-invariant convolution to conventional operations in encoder-decoder network including summation, up-sampling, and concatenation, so that they can implement a novel multi-scale network HMS-Net fused with the RGB feature in multi-scale manners. Li et al. \cite{li2020a} propose to exploit the multi-scale fusion between RGB images and sparse depth samples through cascade hourglass network for depth completion . They extract multi-scale information from image with an encoder and fuse the sparse depth prediction results through the cascaded Hourglass network. The depth sample are downscaled to the corresponding scales as supervisory signals. Hambarde et al. \cite{hambarde2020s2dnet} design a cascaded framework consisting of two convolutional neural networks which are coarse and fine one. The coarse network utilizes a Hourglass network structure to produce coarse prediction. The fine one fuses the coarse prediction and RGB images in multi-scales through attention mechanism. Yang et al. \cite{yang2019dense} design a conditional prior network to extract multi-scale information, which are fused in decoder in multi-scale manners along with valid masks. They exploit the previously seen dataset as prior to estimate the dense map of the RGB images. Inspired by the guided image filtering, Tang et al.  \cite{tang2019learning} develop a guided network to produce the content-dependent, spatially-variant kernels for depth completion. Those kernel weights are automatically learned from RGB images and applied on the sparse depth map in multi-scales. Since the kernels are automatically generated from the content, they are powerful to address challenging scenes. Besides, to reduce GPU memory consumption, a new convolution factorization is implemented with  channel-wise convolution and cross-channel convolution to replace conventional convolutions. Lopez-Rodriguez et al. \cite{lopez-rodriguez2020project} exploit the domain adaptation techniques to address the problem of lacking dense ground truth annotation. They generate the synthetic dataset with driving simulator and add various noise to approximate the real scenes. CycleGAN \cite{CycleGAN2017} is used to transfer real-domain images to the synthetic ones. Wong et al. \cite{wong2020unsupervised} utilize the photometric loss for depth completion . They construct a dense planar scaffolding of the scene with the sparse depth map. Then the scaffolding is refined along with image through skip-connections in multi-scale manner in refinement network. Schuster et al. \cite{2020ssgp} design a  sparse spatial guided propagation method. They predict two affinity matrix from multi-scale information of the RGB images for each scale of the sparse depth map. Then the RGB information is used to refine the predicted coarse depth map following \cite{ChengdepthVA}. Zhao et al. \cite{zhao2020adaptive} extract muti-scale features and propose a co-attention guided graph propagation  strategy for depth completion. The strategy allows to extract the contextual information from RGB images to make compliments of the sparsity patterns of the sparse depth. Besides, they design a symmetric gated fusion strategy to further fuse the multi-modal contextual information.
	
	To alleviate the smooth boundaries in the predicted map, many researchers utilize the low-level features in various manners to constraint the depth prediction.
	Zhang et al. \cite{zhang2018deepcom} conduct the depth completion indirectly in two stages . Firstly, they predict the surface normal and boundary respectively from RGB images. Secondly, the predicted normal map and edge map are globally optimized with the sparse map to get the dense map. Huang et al. \cite{huang2019indoor} utilize the self-attention mechanism to extract features that pay more attention to clear edges. Besides, they also introduce boundary consistency loss based on sobel edge detection results. Lee et al.  \cite{lee2019depth} jointly train two sub-networks: a geometry network and a context network. The geometry network predicts a surface normal of a scene and an initial dense depth map. The context network learns to estimate the bilateral weight, which  captures low level features and global features. The final depth map is generated by multiplying the bilateral weights to the initial dense depth map. Incorporating the low level feature into depth prediction allows to preserve edges and fine details of the depth maps.
	
	Another challenge for depth completion is the noise depth input.  Xu et al.  \cite{xu2019depthcom} exploit the   linear orthogonality between the depth and normal in 3D space to mitigate the effect of the noise input. They first predict the depth and normal jointly in a CNN. Then the predicted depth and normal are transformed into plane-origin distance space and a refinement process is conducted via a diffusion model. Zhang et al.  \cite{zhang2019dfinenet} handle the depth completion problem with noise depth input by jointly predicting the relative pose between two frames with two CNN networks. Both the depth residuals between predicted and ground truth depth map and photometric loss are exploited to train the network. Teixeira et al. \cite{teixeira2020aerial} propose a depth completion for aerial platforms to address large viewpoint and depth variations, and limited computing resources. They perform the depth prediction and confidence prediction by sharing the deep features. 
	
	\section{Applications}\label{sec:app}
	Monocular depth estimation with deep learning techniques has been widely used in SLAM and height estimation with RGB images. 
	
	\subsection{SLAM}
	\subsubsection{Scale recovery}
	Yin et al. \cite{8237887} address scale recovery problem with predicted depth map. The scale can be estimated from the matched points and solved with expectation maximization algorithm. Li et al. \cite{8461251} obtain the scale by training a depth prediction network using sequences of stereo image pairs with photometric loss in spatial and temporal constraints. Wang et al. \cite{8699273} incorporate the focal length parameters into loss to obtain the scale consistency of image depth and the dense map is fused with the key framed depth prediction. Yang et al. \cite{9157454} integrate the depth prediction into visual odometry (VO) system by initializing new 3D points to recover the metric scale. Ye et al. \cite{ye2020drm} perform dense reconstruction by fusing the SLAM output and the CNN-based depth prediction. They scale the predicted depth to the sparse map to keep the scale consist. Wang et al. \cite{wang2019unsupervised} design an unsupervised learning-based method trained with stereo images using temporal constraints. The scale is restored based on the stereo images.
	
	\subsubsection{Map reconstruction}
	Tang et al. \cite{8605349} leverage the benefits of the direct sparse odometry and the CNN-based depth prediction for dense depth prediction. The sparse odometry helps to recover the scale of predicted depth map and depth prediction initializes the direct sparse odometry (DSO) for efficiency. Dense construction is achieved with sparse point clouds and predicted normal as well as the depth map. The final 3D model is refined based on depth prediction results of the key frames. CodeSLAM obtains the dense reconstruction by jointly optimize the learned geometry prior with the camera pose \cite{8578369}. Wang et al. \cite{wang2017deepvo} utilize a virtual stereo view to predict depth. Tateno et al. \cite{8100178} use a CNN to predict the depth and feed it into LSD-SLAM for dense reconstruction and utilize the depth prediction for scale estimation of the dense construction. Ji et al. \cite{8486548} address the dense reconstruction by fusing the CNN-based methods and epipolar-geometric based methods. The CNN-based methods are dense but of blurry edges and the latter are very accurate. The high accurate 3D model can be obtained in this manner. Mukasa et al. \cite{8265321} also produce the 3D reconstruction from the SLAM-measurement and dense predicted depth map from CNNs. They use mesh to represent the 3D model instead of the 3D points and it can be applied for mobile devices. Yang et al. \cite{yang2018deep} incorporate the depth prediction into visual odometry framework to address the scale ambiguity of the video. They train the depth prediction network in a semi-supervised fasion. Laidlow et al. \cite{8793527} address the scale ambiguity of the CNN-based methods by fusing the semi-dense multi view stereo algorithm and the predicted dense depth map and gradients. The scale is recovered by scaling the predicted depth with the focal length of the camera. The gradient is predicted to guarantee the depth consistency. Wang et al. \cite{8699273} present an online monocular dense reconstruction framework using learned depth, which overcomes the inherent difficulties of reconstruction for low-texture regions or pure rotational motions. 
	
	\subsection{Height estimation}
	Ghamisi et al. \cite{8306501} produce digital surface model (DSM) from single aerial images through the conditional generative adversarial networks. The network is trained with paired DSM and image patches with adversarial loss. Mou et al. \cite{mou2018im2height} address the height estimation from the single remote sensing image. They design an encoder-decoder structure based on residual block and train it in an end-to-end manner. The predicted height maps are used for building extraction and better results are achieved. Amirkolaee et al. \cite{amirkolaee2019height}  perform height estimation using convolutional neural networks. They predict the depth of the aerial image patches and merged them with Gaussian smoothing to eliminate the depth jump at the merging area.
	Amirkolaee et al. \cite{amirkolaee2019convolutional}  use convolutional neural network to address digital surface model generation from a single aerial image. They utilize residual block to extract local and global information of each patches and the post-process method is used to connect the estimated DSM patches to generate smooth and continuous surface. 
	Mo et al. \cite{9027856} propose a method to estimate the gradient of the height map instead of directly estimating the height values. The estimated gradients are of two directions and the final elevation is obtained through the maximum likelihood approaches.
	
	\section{Trends and future directions}\label{sec:fut}
	In this section, we will discuss the potential directions of monocular depth prediction, which mainly focus on the accuracy, explainability, and practical deployment issues.
	
	\subsection{Accuracy}
	Although impressive performances have been achieved in monocular depth prediction by previous works, the prediction accuracy is still not comparable with the depth map obtained using distance measuring devices or multi-view geometry. Previous methods based on recurrent neural network, Hourglass Net, GANs, etc., have demonstrated that good network architectures can improve the depth prediction accuracy. Therefore, leveraging new network structures like graph convolution neural network, attention mechanism, etc., may have the potentials to further enhance prediction performance. Exploiting geometric or statistical constraints as regularizers for network training is also a promising direction. For example, introducing the mechanism of blurring and depth consistency of the plane objects as constraints can further force networks to learn the hidden relationship between depth and RGB images. For unsupervised learning-based approaches, there are still plenty room to increase despite the good performance achieved recently. For future research, the first direction is to take fully advantages of geometric or color constraints between adjacent frames or stereo pairs. The second direction is to address the depth scale ambiguity and inconsistency. Currently, most methods try to address it by adding regularization in losses. However, it seems to be a possible solution to tackle the issues in 3D space by aligning point clouds in object level.
	The third direction is to effectively detect dynamic objects and occlusion areas. Current methods address it by learning a mask or through segmentation, one possible solution to it is to fuse segmentation and geometric constraints. 
	
	\subsection{Explainability}
	Deep learning-based depth prediction utilizes convolutional neural networks to learn the depth values from RGB images. Some researchers have tried to answer the question why CNNs are able to infer the depth from images. Hu et al. \cite{9009581} find  that CNNs tend to focus on edges that are related to scene geometry. Besides, they claim that regions near vanishing points also draw significant attentions of the networks. Dijk et al. \cite{9009532} claim that the depth map is estimated based on the vertical position instead of their sizes. Although beneficial attempts have been made, further exploration is still needed and worthy pursuing in this direction.
	
	\subsection{Practical issues}
	Another promising direction is to address the practical issues when putting the deep learning-based methods into practice. There are still two main challenges to be addressed: real-time performance on the embedding devices and transfer-ability.
	Current methods usually rely on heavily parameterized deep neural networks to achieve high accuracy, which means large computation and storage requirement. It works on powerful computers, however is infeasible for embedding devices.  Designing lightweight networks for practical applications is a very promising research direction. Recent works have shown that lightweight networks can achieve competitive performances with much smaller number of parameters and only slight performance drop \cite{8594243}. Therefore, it is worth trying new convolutional operations like depth-wise separable convolution, group convolution, or dilated convolution to modify current network structures. In addition, deep neural networks often suffer from the over-fitting problem. Deep learning-based depth prediction also have such a problem. Current methods work well when trained and tested using images captured from the same cameras in similar scenes. However, in real-world scenes, the images are often taken in various scenes using various cameras, which would lead to significant performance drop of current methods. To alleviate these issues, online learning or domain adaptation techniques can be exploited in future work.
	
	\section{Concluding remarks} \label{sec:con}
	In this paper, we provide a thorough review of the recent works on deep learning-based monocular depth prediction. We summarize the recent works from three perspectives: supervised-based methods, unsupervised-based methods, and depth completion. The evolution of network structure and loss are elaborated for supervised methods. We over-view the unsupervised-based methods based on the key questions. Depth completion is reviewed from the perspective of fusion and information extraction facts. We also point out the future directions. Such a review is necessary for monocular depth prediction since this field has been received increasing interests. With this review, we can identify where we are and which directions to follow.
	
	\section*{Acknowledgments}
	This work was supported in part by the National Key R\&D Program of China 2018YFB2101000, in part by the National Natural Science Foundation of China under Grant 41871329, in part by the Science and Technology Planning Project of Guangdong Province under Grant 2018B020207005.
	
	\bibliographystyle{numcompress}
	\bibliography{depthreview}
	
\end{document}